\documentclass{ecai2014}
\usepackage{times}
\usepackage{latexsym}
\usepackage{url}
\usepackage{xspace}
\usepackage{amsmath}
\usepackage{amssymb}

\newcommand{\mi}[1]{\ensuremath{\mathit{#1}}}

\newcommand{\define}[1]{\emph{#1}}

\newcommand{\la}{\leftarrow}

\newcommand{\bstate}[0]{\ensuremath{\mathsf{S}}\xspace}

\newcommand{\op}[1]{\ensuremath{\mathsf{#1}}\xspace}

\newcommand{\attime}[2]{\ensuremath{#1^{#2}}}
\newcommand{\timesensor}[0]{\ensuremath{\mathsf{t}}}

\newcommand{\Contexts}[0]{\ensuremath{\mathsf{C}}\xspace}
\newcommand{\BRs}[0]{\ensuremath{\mathsf{BR}}\xspace}
\newcommand{\KBs}[0]{\ensuremath{\mathsf{KB}}\xspace}

\newcommand{\lit}[2]{\ensuremath{\pred{#1}(\term{#2})}\xspace}
\newcommand{\pred}[1]{\ensuremath{\mathrm{#1}}\xspace}
\newcommand{\term}[1]{\ensuremath{\mathit{#1}}\xspace}
\newcommand{\brc}[2]{\ensuremath{{#1}{:}{#2}}\xspace}
\newcommand{\brclit}[3]{\ensuremath{#1{:}\lit{#2}{#3}}\xspace}
\newcommand{\brcpred}[2]{\ensuremath{#1{:}\pred{#2}}\xspace}
\newcommand{\bro}[2]{\ensuremath{{#2}{@}{#1}}\xspace}
\newcommand{\brolit}[3]{\ensuremath{\bro{#1}{\lit{#2}{#3}}}\xspace}
\newcommand{\bropred}[2]{\ensuremath{\bro{#1}{\pred{#2}}}\xspace}

\newcommand{\head}[1]{\ensuremath{\mi{hd}({#1})}\xspace}
\newcommand{\body}[1]{\ensuremath{\mathrm{bd}({#1})}\xspace}

\newcommand{\naf}[0]{\ensuremath{\mathrm{not}}\xspace}

\newcommand{\tuple}[1]{\ensuremath{\tupleLeft {#1} \tupleRight}\xspace}
\newcommand{\tupleLeft}[0]{\ensuremath{\langle}\xspace}
\newcommand{\tupleRight}[0]{\ensuremath{\rangle}\xspace}
\newcommand{\set}[1]{\{#1\}}

\newcommand{\etal}[0]{\emph{et al.}\xspace}
\newcommand{\wrt}{wrt.\xspace}

\newcommand{\iec}[0]{i.e.,\xspace}

\newcommand{\egc}[0]{e.g.,\xspace}

\newcommand{\bi}[0]{\begin{itemize}}
\newcommand{\ei}[0]{\end{itemize}}

\newtheorem{definition}{Definition}
\newtheorem{example}{Example}

\newcommand{\problemA}[0]{\ensuremath{Q^\forall}\xspace}
\newcommand{\problemE}[0]{\ensuremath{Q^\exists}\xspace}

\begin{document}

\title{Multi-Context Systems for \\ Reactive Reasoning in Dynamic Environments\thanks{A former version~\cite{Brewka2014} of this paper has been accepted for publication in the Procs. of ECAI-14. This work has been partially supported by the German Research Foundation (DFG) under grants BR-1817/7-1 and FOR 1513.}}

\author{Gerhard Brewka \and Stefan Ellmauthaler \and J\"org P\"uhrer\institute{Institute of Computer Science, Leipzig University,
Germany, email: \{brewka,ellmauthaler,puehrer\}@informatik.uni-leipzig.de} }

\maketitle
\bibliographystyle{ecai2014}

\begin{abstract}
We show in this paper how managed multi-context systems (mMCSs) can be turned into a reactive formalism suitable for continuous reasoning in dynamic environments. We extend mMCSs with (abstract) sensors and define the notion of a run of the extended systems. We then show how typical problems arising in online reasoning can be addressed: handling potentially inconsistent sensor input, modeling intelligent forms of forgetting, selective integration of knowledge, and controlling the reasoning effort spent by contexts, like setting contexts to an idle mode. We also investigate the complexity of some important related decision problems and discuss different design choices which are given to the knowledge engineer.

\end{abstract}

\section{Introduction}
Research in knowledge representation (KR) faces two major problems. First of all,
a large variety of different
 languages for representing knowledge - each of them useful for particular types of problems - has been produced. There are many situations where the integration of the knowledge represented in
diverse formalisms is crucial, and principled ways of achieving this integration are needed. Secondly,
most of the tools providing reasoning services for KR languages were developed for offline usage: given a knowledge base (KB) computation is one-shot, triggered by a user, through a specific query or a request to compute, say, an answer set. This is the right thing for specific types of applications where a specific answer to a particular problem instance is needed at a particular point in time. However, there are different kinds of applications where a reasoning system is continuously online and receives information about a particular system it observes.
 Consider an assisted living scenario where people in need of support live in an apartment equipped with various sensors, \egc smoke detectors, cameras, and body sensors measuring relevant body functions (\egc pulse, blood pressure). A reasoning system continuously receives sensor information. The task is to detect emergencies (health problems, forgotten medication, overheating stove,...) and cause adequate reactions (\egc turning off the electricity, calling the ambulance, ringing an alarm). The system is continuously online and has to process a continuous stream of information rather than a fixed KB.

This poses new challenges on KR formalisms. Most importantly, the available information continuously grows.
This obviously cannot go on forever as the KB needs to be kept in a manageable size. We thus need principled ways of forgetting/disregarding information. In the literature one often finds sliding window techniques~\cite{Gebser2012} where information is kept for a specific, predefined period of time  and forgotten if it falls out of this time window. We believe this approach is far too inflexible. What is needed is a dynamic, situation dependent way of determining whether information needs to be kept or can be given up. Ideally we would like our online KR system to guarantee specific response times; although it may be very difficult to come up with such guarantees, it is certainly necessary to find means to identify and focus on relevant parts of the available information.
Moreover, although the definition of the semantics of the underlying KR formalism remains essential, we also need to impose procedural aspects reflecting the necessary modifications of the KB. This leads to a new, additional focus on \emph{runs} of the system, rather than single evaluations.

Nonmonotonic multi-context systems (MCSs)
\cite{BrewkaE07}
 were explicitly developed to handle the integration problem. In a nutshell, an MCS consists of reasoning units - called contexts for historical reasons \cite{GiunchigliaS94} - where each unit can be connected with other units via so-called bridge rules. The collection of bridge rules associated with a context specifies additional beliefs the context is willing to accept depending on what is believed by connected contexts. The semantics of the MCS is then defined in terms of equilibria. Intuitively, an equilibrium is a collection of belief sets, one for each context, which fit together in the sense that the beliefs of each context adequately take into account what the other contexts believe.

The original framework was aimed at modeling the flow of information among contexts, consequently the addition of information to a context was the only possible operation on KBs. To capture more general forms of operations MCSs were later generalized to so called managed MCSs (mMCSs) \cite{BrewkaEFW11}. The main goal of this paper is to demonstrate that this additional functionality makes managed MCSs particularly well-suited as a basis for handling the mentioned problems of online reasoning systems as well. The main reason is that the operations on the knowledge bases allow us to control things like KB size, handling of inconsistent observations, focus of attention, and even whether a particular context should be idle for some time.

However, to turn mMCSs into a reactive online formalism we first need to extend the framework to accommodate observations. We will do so by generalizing bridge rules so that they have access not only to belief sets of other contexts, but also to sensor data. This allows systems to become \emph{reactive}, that is, to take information about a dynamically changing world into account and to modify themselves to keep system performance up.

The rest of the paper is organized as follows. We first give the necessary background on mMCSs. We then extend the framework to make it suitable for dynamic environments, in particular we show how observations can be accommodated, and we define the notion of a run of an MCS based on a sequence of observations. The subsequent sections address the following issues: handling time and the frame problem; dynamic control of KB size; focus of attention; control of computation (idle contexts). We finally discuss the complexity of some important decision problems.\footnote{The paper is based on preliminary ideas described in the extended abstract \cite{Brewka2013a} and in \cite{Ellmauthaler2013a}. However, the modeling techniques as well as the formalization presented here are new. A key difference in this respect is the handling of sensor data by means of bridge rules.}

\section{Background: Multi-Context Systems}

We now give the necessary background on managed MCSs \cite{BrewkaEFW11} which provides the basis for our paper. We present a slightly simplified variant of mMCSs here as this allows us to better highlight the issues relevant for this paper. However, if needed it is rather straightforward (albeit technically somewhat involved) to extend all our results to the full version of mMCSs. More specifically we make two restrictions: 1) we assume all contexts have a single logic rather than a logic suite as in \cite{BrewkaEFW11}; 2) we assume that management functions are deterministic.

In addition we will slightly rearrange the components of an mMCS which makes them easier to use for our purposes. In particular, we will keep bridge rules and knowledge bases separate from their associated contexts. The latter will change dynamically during updates, as we will see later, and it is thus convenient to keep them separate.
Bridge rules will be separated due to technical reasons (\iec better presentation of the later introduced notion of a run).

An mMCS builds on an abstract notion of a \emph{logic} $L$ as a
triple $(\mathit{KB}_L,\mathit{BS}_L, \mathit{ACC}_L)$, where
$\mathit{KB}_L$ is the set of
admissible knowledge bases (KBs) of $L$, which are sets of KB-elements
(``formulas''); $\mathit{BS}_L$ is the set of possible belief sets,
whose elements are beliefs; and
$\mathit{ACC}_L : \mathit{KB}_L \to 2^{\mathit{BS}_L}$ is a function describing the
semantics of $L$ by assigning to each KB a set of acceptable
belief sets.

\begin{definition}
A context is of the form $C=\tuple{L,ops, mng}$ where
\begin{itemize}
  \item $L$ is a logic,
  \item $ops$ is a set of operations,
  \item $mng:2^{ops} \times \mathit{KB}_{L} \rightarrow \mathit{KB}_{L} $ is a management function.
  \end{itemize}
\end{definition}
\noindent
For an indexed context $C_i$ we will write $L_i$, $ops_i$, and $mng_i$
to denote its components.

\begin{definition}\label{def:br}
Let $\Contexts = \tuple{C_1,\ldots,C_n}$ be a tuple of contexts. A bridge rule for $C_i$ over $\Contexts$ ($1 \leq i \leq n$) is of the form \begin{align}
  \label{bridgerule}
    \op{op} \la& a_1,\ldots, a_j,\naf\ a_{j+1},\ldots,\naf\ a_m,
\end{align}
such that $\op{op} \in ops_i$ 
and every $a_\ell$ ($1 \leq \ell \leq m$) is an atom of form \brcpred{c}{b}, where 
$c\,{\in}\, \{1,\ldots,n\}$, and $\pred{b}$ is a belief for $C_c$, \iec $\pred{b}\in S$ for some $S\in\mathit{BS}_{L_c}$.
\end{definition}
For a bridge rule $r$, the operation $\head{r}=\op{op}$ is the \define{head} of $r$,
while $\body{r}=\{a_1,\ldots, a_j,\naf\ a_{j+1},\ldots,\naf\ a_m\}$
is the \define{body} of $r$.

\begin{definition}\label{def:mcs}
A \emph{managed multi-context system (mMCS)}\/ $M=\tuple{\Contexts,\BRs,\KBs}$ is a triple consisting of \begin{enumerate} \item a tuple of contexts $\Contexts=\tuple{C_1,\ldots,C_n}$,
\item a tuple $\BRs =\tuple{br_1,\ldots,br_n}$, where each $br_i$ is a set of bridge rules for $C_i$ over $\Contexts$,

\item
a tuple of KBs $\KBs =\tuple{kb_1,\ldots,kb_n}$ such that $kb_i \in \mathit{KB}_{L_i}$. \end{enumerate}
\end{definition}

A belief state $\bstate=\tuple{S_1,\ldots,S_n}$ for $M$ consists of belief sets
$S_i \in \mathit{BS}_{L_i}$, $1\leq i \leq n$.
Given a bridge rule $r$,
an atom $\brcpred{c}{p}\in\body{r}$ is satisfied by $\bstate$ if $\pred{p} \in S_{c}$ and
a negated atom $\naf\ \brcpred{c}{p}\in\body{r}$ is satisfied by $\bstate$ if $\pred{p} \not\in S_{c}$.
A literal is an atom or a negated atom.
We say that $r$ is applicable wrt.~$\bstate$, denoted by $\bstate \models
\body{r}$, if every literal $l\in\body{r}$ is satisfied by $\bstate$.
We use $\mi{app}_i(\bstate) = \{ \head{r} \mid r \in br_i \land \bstate \models
\body{r}\}$  to denote the heads of all applicable bridge
rules of context $C_i$ wrt.~$\bstate$.

The semantics of an mMCS $M$ is then defined in terms of equilibria, where an \emph{equilibrium}
is a belief state $\bstate=\tuple{S_1,\ldots,S_n}$ satisfying the following condition: the belief set chosen for each context must be acceptable for the  KBs obtained by applying the management function to the heads of applicable bridge rules and the KB associated with the context.
More formally, for all contexts $C_i = \tuple{L_i, ops_i, mng_i}$:
    let $S_i$ be the belief set chosen for $C_i$.
Then $\bstate$ is an equilibrium if, for $1 \leq i \leq n$, $$S_i \in ACC_i(kb') \mbox{ for } kb' = mng_i(\mi{app}_i(\bstate),kb_i).$$
Management functions allow us to model all sorts of modifications of a context's KB and thus make mMCSs a powerful tool for describing the influence contexts can have on each other.

\section{Reactive Multi-Context Systems}\label{sec:rmcs}

To make an mMCS $M$ suitable for reactive reasoning in dynamic environments, we have to accomplish two tasks: \begin{enumerate}
\item we must provide means for the MCS to obtain information provided by sensors, and
\item we have to formally describe the behavior of the MCS over time.
\end{enumerate}

Let us first show how sensors can be modeled abstractly. We assume that a sensor $\Pi$ is a device which is able to provide new information in a given language $L_{\Pi}$ specific to the sensor. From an abstract point of view, we can identify a sensor with its observation language and a current sensor reading, that is, $\Pi = \tuple{L_{\Pi},\pi}$ where $\pi \subseteq L_{\Pi}$. Given a tuple of sensors $\Pi = \tuple{\Pi_1, \ldots, \Pi_k}$, an observation $Obs$ for $\Pi$ ($\Pi$-observation for short) consists of a sensor reading for each sensor, that is, $Obs = \tuple{\pi_1, \ldots, \pi_k}$ where for $1 \leq i \leq k$, $\pi_i \subseteq L_{\Pi_i}$.

Each context must have access to its relevant sensors. Contexts already have means to obtain information from outside, namely the bridge rules.  This suggests that the simplest way to integrate sensors is via an extension of the bridge rules: we will assume that bridge rules in their bodies can not only refer to contexts, but also to sensors.

\begin{definition}
A reactive multi-context system (rMCS) over sensors $\Pi = \tuple{\Pi_1, \ldots, \Pi_k}$ is a tuple $M = \tuple{\Contexts, \BRs, \KBs}$,
as in Def.~\ref{def:mcs} except that
the atoms $a_\ell$ ($1 \leq \ell \leq m$) of bridge rules in $\BRs$ for context $C_i$ of form~(\ref{bridgerule})
can either be a \define{context atom} of form $\brcpred{c}{b}$ as in Def.~\ref{def:br},
or a \define{sensor atom} of form $\bropred{s}{o}$, where $s$ is an index determining a sensor ($1\leq s\leq k$)
and $\pred{o}\in L_{\Pi_s}$ is a piece of sensor data.
\end{definition}
\noindent
The applicability of bridge rules
now also depends on an observation:
\begin{definition}
Let $\Pi$ be a tuple of sensors and $Obs = \tuple{\pi_1, \ldots, \pi_k}$ a $\Pi$-observation.
A sensor atom $\bropred{s}{o}$  is \define{satisfied} by $Obs$ if $o \in \pi_s$; a literal $\naf\ \bropred{s}{o}$ is \define{satisfied} by $Obs$ if $o \not \in \pi_s$.

Let $M = \tuple{\Contexts, \BRs, \KBs}$ be an rMCS with sensors $\Pi$ and
$\bstate$ a belief state for $M$.
A bridge rule $r$ in $\BRs$ is \define{applicable} wrt. $\bstate$ and $Obs$, symbolically $\bstate\models_{Obs}\body{r}$, if
every context literal in \body{r} is satisfied by $\bstate$ and every sensor literal in \body{r} is satisfied by $Obs$.
\end{definition}
Instead of $\mi{app}_i(\bstate)$ we use $\mi{app}_i(\bstate,Obs) = \{ \head{r} \mid r \in br_i \land \bstate\models_{Obs}\body{r}\}$
to define an equilibrium of an rMCS in a similar way as for an mMCS:
\begin{definition}
Let $M = \tuple{\Contexts, \BRs, \KBs}$ be an rMCS with sensors $\Pi$ and $Obs$ a $\Pi$-observation.
A belief state  $\bstate = \tuple{S_1, \ldots, S_n}$ for $M$ is an equilibrium of $M$ under $Obs$ if, for $1 \leq i \leq n$,
$$S_i \in ACC_i(mng_i(\mi{app}_i(\bstate,Obs),kb_i)).$$
\end{definition}

\begin{definition}
Let $M = \tuple{\Contexts, \BRs, \KBs}$ be an rMCS with sensors $\Pi$, $Obs$ a $\Pi$-observation, and $\bstate = \tuple{S_1, \ldots, S_n}$ an equilibrium of $M$ under $Obs$. The tuple of KBs generated by $\bstate$ is defined as 
$\KBs^\bstate = \tuple{mng_1(\mi{app}_1(\bstate,Obs),kb_1), \ldots, mng_n(\mi{app}_n(\bstate,Obs),kb_n)}$.
The pair $\tuple{\bstate,\KBs^\bstate}$ is called full equilibrium of $M$ under $Obs$.
\end{definition}
We now introduce the notion of a run of an rMCS induced by a sequence of observations:

\begin{definition}
Let $M = \tuple{\Contexts, \BRs, \KBs}$  be an rMCS with sensors $\Pi$ and $O = (Obs^0, Obs^1, \ldots)$ a sequence of $\Pi$-observations. A run of $M$ induced by $O$ is a sequence of pairs $R = (\tuple{\bstate^0,\KBs^0}, \tuple{\bstate^1,\KBs^1}, \ldots)$ such that
\begin{itemize}
\item $\tuple{\bstate^0,\KBs^0}$ is a full equilibrium of $M$ under $Obs^0$,
\item for $\tuple{\bstate^i,\KBs^i}$ with $i > 0$, $\tuple{\bstate^i,\KBs^i}$ is a full equilibrium of $\tuple{C,\BRs, \KBs^{i-1}}$ under $Obs^i$.
\end{itemize}
\end{definition}

To illustrate the notion of a run, let's discuss a simple example. We want to model a clock which allows other contexts to add time stamps to sensor information they receive. We consider two options. We will first show how a clock can be realized which generates time internally by increasing a counter whenever a new equilibrium is computed. We later discuss a clock based on a sensor having access to ``objective" time. In both cases we use integers as time stamps.
\begin{example}\label{ex:clock}
Consider a context $C_c$ whose KBs (and belief sets) are of the form $\{\lit{now}{t}\}$ for some integer $t$. Let $kb^0 = \{\lit{now}{0}\}$. Assume the single bridge rule of the context is $\op{incr} \la$, which intuitively says time should be incremented whenever an equilibrium is computed. The management function is thus defined as $$mng_c(\{\op{incr}\},\{\lit{now}{t}\}) = \{\lit{now}{t+1}\}$$ for each $t$. Since the computation of the (full) equilibrium is independent of any other contexts and observations, the context just increments its current time whenever a new equilibrium is computed. Each run of an rMCS with context $C_c$  will thus contain for $C_c$ the sequence of belief sets $\{\lit{now}{1}\}$, $\{\lit{now}{2}\}$, $\{\lit{now}{3}\}, \dots$. The example illustrates that the system may evolve over time even if there is no observation made at all.

It is illustrative to compare this with a context $C_{c'}$ which is like the one we discussed except for the bridge rules which now are the instances of the schema $$\op{set(\lit{now}{T+1})} \la \brclit{c'}{now}{T}.$$ The management function correspondingly becomes $$mng_{c'}(\{set(\lit{now}{t+1})\},\{\lit{now}{t}\}) = \{\lit{now}{t+1}\}$$ for all $t$. Note that in this case no equilibrium exists! The reason for this is that by replacing $\lit{now}{0}$ with $\lit{now}{1}$ the precondition for the rule sanctioning this operation goes away. Special care thus needs to be taken when defining the operations.
\end{example}

In the rest of the paper we will often use an alternative approach where ``objective" time is entered into the system by a particular sensor $\Pi_\timesensor$. In this case each update of the system makes time available to each context via the current sensor reading of $\Pi_\timesensor$.

In Example~\ref{ex:clock} we already used a bridge rule schema, that is, a bridge rule where some of the parts are described by parameters (denoted by uppercase letters). We admit such schemata to allow for more compact representations. A bridge rule schema is just a convenient abbreviation for the set of its ground instances. The ground instances are obtained by replacing parameters by adequate ground terms. We will admit parameters for integers representing time, but also for formulas and even contexts. In most cases it will be clear from the discussion what the ground instances are, in other cases we will define them explicitly. We will also admit some kind of basic arithmetic in the bridge rules and assume the operations to be handled by grounding, as is usual, say, in answer set programming. For instance,  the bridge rule schema $$\op{add(\lit{p}{T+1})} \la \brclit{c}{p}{T}, \naf\ \brclit{c}{\neg p}{T+1}$$ which we will use to handle the frame problem in the next section has ground instances $\op{add(\lit{p}{1})} \la \brclit{c}{p}{0}, \naf\ \brclit{c}{\neg p}{1}$, $\op{add(\lit{p}{2})} \la \brclit{c}{p}{1}, \naf\ \brclit{c}{\neg p}{2}$, etc.

Although in principle parameters for integers lead to an infinite set of ground instances, in our applications only ground instances up to the current time (or current time plus a small constant, see Sect.~\ref{sec:control}) are needed, so the instantiations of time points remain finite.

In the upcoming sections we describe different generic modeling techniques for rMCSs.
For concrete applications, these techniques can be refined and tailored
towards the specific needs of the problem domain at hand.
To demonstrate this aspect, we provide a more specific example from an assisted living application.
\begin{example}\label{ex:bob1}
Although Bob suffers from dementia, he is able to live
in his own apartment as it is equipped with an assisted living system
that we model by means of an rMCS.
Assume Bob starts to prepare his meal.
He leaves the kitchen to go to the bathroom.
After that, he forgets he has been cooking,
goes to bed and falls asleep.
The rMCS should be able to recognize a potential emergency
based on the data of different sensors in the flat that monitor, \egc the state of the kitchen
equipment and track Bob's position.

Our rMCS $M$ 
has three contexts $\Contexts=\tuple{C_{kt},C_{hu},C_{ig}}$ and sensors $\Pi=\tuple{\Pi_{pow},\Pi_{tmp},\Pi_{pos}}$.
$C_{kt}$ 
is the kitchen equipment context that monitors Bob's stove.
Its formulas and beliefs are
atoms from  $at_{kt}=\set{\lit{pw}{on},\lit{pw}{off},\lit{tm}{cold},\lit{tm}{hot}}$  representing the stove's power status (on/off)
and a qualitative value for its temperature (cold/hot).
The logic 
 $L_{kt}=\tuple{2^{at_{kt}},2^{at_{kt}},ACC_{id}}$ of $C_{kt}$ 
has a very simple semantics $ACC_{id}$ in which every knowledge base $kb$
has only one accepted belief set coinciding with the formulas of $kb$, \iec
$ACC_{id}(kb)=\set{kb}$.
The bridge rules for $C_{kt}$ over $\Contexts$ are 
$$
\begin{array}{r@{}l}
\op{setPower(\term{P})} \la &\brolit{pow}{switch}{P}.\\
\op{setTemp(\term{cold})} \la &\bro{tmp}{T},T\leq 45.\\
\op{setTemp(\term{hot})} \la &\bro{tmp}{T}, 45 < T.
\end{array}
$$
that react to switching the stove on or off, registered by sensor $\Pi_{pow}$,
respectively read numerical temperature values from sensor $\Pi_{tmp}$ and
classify the temperature value as cold or hot.
The management function $mng_{kt}(app,kb)=$
$$
\begin{array}{l@{}l}
\{\lit{pw}{on}\mid& \op{setPower(\term{on})}\in app \lor \\&(\lit{pw}{on}\in kb \land \op{setPower(\term{off})}\not\in app) \}\cup\\
                 \{\lit{pw}{off}\mid& \op{setPower(\term{on})}\not\in app \land\\&
                  (\lit{pw}{on}\not\in kb \lor  \op{setPower(\term{off})}\in app)\}\cup\\
                 \{\lit{tm}{t}\mid& \op{setTemp(\pred{t})}\in app\}
\end{array}
$$
ensures that the stove is considered on when it is switched on or when it is not being switched off and already considered on in
the old knowledge base $kb$. Otherwise, the KB constructed by the management function contains the atom \lit{pw}{off}.
Context $C_{hu}$ keeps track of Bob's position.
The language of sensor $\Pi_{pos}$ is given by $L_{\Pi_{pos}}=\set{\lit{enters}{kitchen},\lit{enters}{bathroom},\lit{enters}{bedroom}}$
and non-empty sensor readings of $\Pi_{pos}$ signal when Bob has changed rooms.
The semantics of $C_{hu}$ is also $ACC_{id}$ and
its bridge rules are given by the schema
$$\op{setPos(\term{P})} \la \brolit{pos}{enters}{P}.$$
The management function writes Bob's new position into the KB whenever he changes rooms and keeps the previous
position, otherwise.
$C_{ig}=\tuple{L_{ig},ops_i, mng_{ig}}$ is the context for detecting emergencies.
It is implemented as an answer-set program, hence the acceptable belief sets of $L_{ig}$ are the answer sets
of its KBs.
The bridge rules of $C_{ig}$ do not refer to sensor data but query other contexts:
$$
\begin{array}{l}
\op{extVal(\lit{oven}{P,T})} \la \brclit{kt}{pw}{P},\brclit{kt}{tm}{T}.\\
\op{extVal(\lit{humanPos}{P})} \la \brclit{hu}{pos}{P}.\\
\end{array}
$$
The answer-set program $kb_{ig}$ is given by the rule
$$
\begin{array}{l}
\pred{emergency} \la \lit{oven}{on,hot}, \naf\ \lit{humanPos}{kitchen}.\\
\end{array}
$$
The management function of $C_{ig}$ that adds information from the bridge rules temporarily
as input facts to the context's KB is given by $mng_{ig}(app,kb)=$
$$
\begin{array}{l@{}l}
(kb \setminus & (\set{\lit{oven}{P,T}\la\mid P\in\set{\term{on},\term{off}},T\in\set{\term{cold},\term{hot}}}\cup\\&
				 \set{\lit{humanPos}{R}\la\mid \lit{enters}{R}\in L_{\Pi_{pos}}}))\cup\\
\multicolumn{2}{l}{
\set{\lit{oven}{\term{p},\term{t}}\la\mid \op{extVal(\lit{oven}{p,t})\in app}}\cup
}\\
\multicolumn{2}{l}{
\set{\lit{humanPos}{\term{r}}\la\mid \op{extVal(\lit{humanPos}{r})}\in app}.
}
\end{array}
$$
Consider the sequence
$O=(\attime{Obs}{0},\attime{Obs}{1})$
of $\Pi$-observations with
$
\attime{Obs}{i}=\tuple{\attime{\pi_{pow}}{i},\attime{\pi_{tmp}}{i},\attime{\pi_{pos}}{i}}
$
for $0\leq i \leq 1$,
$\attime{\pi_{pow}}{0}=\{\lit{switch}{on}\}$,
$\attime{\pi_{tmp}}{0}=\{16\}$, $\attime{\pi_{tmp}}{1}=\{81\}$,
$\attime{\pi_{pos}}{0}=\{enters(kitchen)\}$, $\attime{\pi_{pos}}{1}=\{enters(bathroom)\}$,
and $\attime{\pi_{s}}{i}=\emptyset$ for all other $\attime{\pi_{s}}{i}$.
Then, $\tuple{\attime{\bstate}{0},\attime{\KBs}{0}}$ is a full equilibrium of $M$ under $\attime{Obs}{0}$, where
$$
\begin{array}{l@{}l}
\attime{\bstate}{0}=\tupleLeft&\set{\lit{pw}{on},\lit{tm}{cold}}, \set{\lit{pos}{kitchen}},\\&\set{\lit{oven}{on,cold},\lit{humanPos}{kitchen}}\tupleRight.
\end{array}
$$
and $\attime{\KBs}{0}$ equals $\attime{\bstate}{0}$ except for the last component which is
$kb_{ig}\cup\set{\lit{oven}{on,cold}\la,\lit{humanPos}{kitchen}\la}$.
Moreover, $(\tuple{\attime{\bstate}{0},\attime{\KBs}{0}}, \tuple{\attime{\bstate}{1},\attime{\KBs}{1}})$
is a run of $M$ induced by $O$, where
$$
\begin{array}{l@{}l}
\attime{\bstate}{1}=\tupleLeft&\set{\lit{pw}{on},\lit{tm}{hot}}, \set{\lit{pos}{bathroom}},\\&\set{\lit{oven}{on,hot},\lit{humanPos}{bathroom},\pred{emergency}}\tupleRight.
\end{array}
$$
\end{example}

\section{Handling sensor data}\label{sec:data}

In this section we discuss how to model an rMCS where possibly inconsistent sensor data can be integrated into a context $C_j$.
To this end, we add a time tag to the sensor information and base our treatment of time on the second option discussed in the last section, that is, we assume a specific time sensor $\Pi_\timesensor$ that yields a reading $\pi_\timesensor$ of the actual time of the form $\lit{now}{t}$ where $t$ is an integer.

Let $\Pi_{j_1}, \ldots, \Pi_{j_m}$ be the sensors which provide relevant information for $C_j$ in addition to $\Pi_\timesensor$. Then $C_j$ will have bridge rules of the form $$\op{add(\pred{P},\term{T},\mathit{j_r})} \la \bropred{j_r}{P}, \bropred{\timesensor}{\lit{now}{T}}$$ where the operation $\op{add}$ is meant to add new, time tagged information to the context.

We assume the readings of a single sensor at a particular time point to be consistent. However, it is a common problem that the readings of different sensors may be inconsistent with each other \wrt some context-dependent notion of inconsistency. To handle this we foresee a management function $mng_j$ that operates based on a total preference ranking of the available sensors. The third argument of the $\op{add}$ operation provides information about the source of sensor information and thus a way of imposing preferences on the information to be added. Without loss of generality assume $j_1 > \ldots > j_m$, that is, sensor $\Pi_{j_1}$ has highest priority.

Now let $add(\bstate)$ be the set of add-operations in the heads of bridge rules active in belief state $\bstate$. We define $$Add_{j_1}(\bstate) = \{(\pred{p},t) \mid \op{add(\pred{p},\mathit{t},\mathit{j_1})} \in add(\bstate)\}$$ and for $1 < i \leq m$ we let $Add_{j_{i}}(\bstate) = Add_{j_{i-1}}(\bstate) \cup$ $$ \{(\pred{p},t) \mid \op{add(\pred{p},\mathit{t},\mathit{j_{i}})} \in add(\bstate), (\pred{p},t) \mbox{ consistent with } Add_{j_{i-1}}(\bstate)\}.$$ Finally, we define $mng_j(add(\bstate),kb) = kb \cup Add_{j_m}(\bstate)$.

This shows how the management function can solve conflicts among inconsistent sensor readings based on preferences among the sensors. Of course, many more strategies of integrating inconsistent sensor data can be thought of which we are not going to discuss in the paper. Please also note that the bridge rules do not necessarily have to pass on sensor information as is to the context. They may as well provide the context with some abstraction of the actual readings. For instance, the sensor temperature information $temp = 55$ may be transformed into qualitative information by a rule schema like
$$
\begin{array}{l@{}l}
\op{add(\mathit{temp = high},\term{T},\mathit{j_r})} \la &\bro{j_r}{temp = x}, 45 \leq x \leq 65,\\& \brolit{\timesensor}{now}{T}.
\end{array}
$$

We next present a way to address the frame problem using bridge rules
when sensors are not guaranteed to provide complete information about the state of the environment in each step. 
In this case we want to assume, at least for some of the atoms or literals observed at time $T-1$ which we call persistent, that they also hold at time $T$.

Assume $\pred{p}$ is some persistent observable property. Persistence of $\pred{p}$ is achieved by the following bridge rule schema:
$$\op{add(\lit{p}{T})} \la \brolit{\timesensor}{now}{T}, \brclit{j}{p}{T-1}, \naf\ \brclit{j}{\neg p}{T}.$$

Please note that in order to avoid non-existence of equilibria as discussed at the end of Sect.~\ref{sec:rmcs} the use of this rule schema for the frame problem presupposes that information about $\pred{p}$ valid at time $T-1$ remains available and is not deleted by any other bridge rule.

\section{Selective forgetting and data retention}\label{sec:forget}

To illustrate our approach we discuss in this section a context $C_d$ which can be used for emergency detection in dynamic environments. Assume there are $m$ potential emergencies $E_1, \ldots, E_m$ we want the context to handle. The role of $C_d$ is to check, based on the observations made, whether one or more of the emergencies $E_i$ are suspected or confirmed. Based on information about potential emergencies $C_d$ adjusts the time span observations are kept. This is the basis for intelligent forgetting based on dynamic windows.

We do not make any assumption about how $C_d$ works internally apart from the following:

\begin{itemize}
  \item $C_d$ may signal that emergency $E_i$ is suspected ($susp(E_i)$) or confirmed ($conf(E_i)$).
  \item $C_d$ has information about default, respectively actual window sizes for different observations ($def.win(p,x)$, $win(p,x)$), and
  \item about the number of time steps observations are relevant for particular emergencies ($rel(p,e,x)$).
\end{itemize}
Given facts of the form mentioned above, here is a possible collection of bridge rules for the task. The operation $set$ sets the window size to a new value, deleting the old one.
To signal an alarm, information is added to the context KB via the operation $alarm$.
\begin{tabbing}
xxxxx \= xxxxxxxxxxxxx \= xxi \= \kill
\> $\op{set(\lit{win}{P,X})}$ \> $\la$ \> $\brclit{d}{def.win}{P,X}, \naf\ \brclit{d}{susp}{E}$ \\
\> $\op{set(\lit{win}{P,Y})}$ \> $\la$ \> $\brclit{d}{rel}{P,E,Y}, \brclit{d}{susp}{E}$ \\
\> $\op{alarm(\term{E})}$ \> $\la$ \> $\brclit{d}{conf}{E}$
\end{tabbing}

Finally, we have to make sure deletions of observations are performed in accordance with the determined window sizes:
$$\op{del(\lit{p}{T'})} \la \brolit{\timesensor}{now}{T}, \brclit{d}{win}{P,Z}, T' < T-Z.$$

The management function just performs additions and deletions on the context KB. Since additions always are tagged with the current time, whereas deletions always refer to an earlier time, there can never be a conflict.

We have so far described a form of focusing where a time window is extended based on a specific suspected event. The next example shows a different form of focusing where specific information is generated and kept only during there is a potential danger in a particular room.

\begin{example}\label{ex:bob2}
Continuing Example~\ref{ex:bob1} we show how context $C_{ig}$ can focus on specific rooms if there is a potential emergency.
For the kitchen there is a threat if the stove is on,
and it then becomes important to track whether someone is in the kitchen. Assume $C_{ig}$ has a potential belief $\lit{pw}{on,T}$ expressing the stove is $on$ since $T$. Focusing on the kitchen can be modeled by following the ASP-rule in $C_{ig}$'s KB:
$$
\lit{focus}{kitchen} \la \lit{pw}{on,T}.
$$
In addition we will need a bridge rule, which keeps track whether Bob is absent from a room in case that room is in the current focus:
\begin{align*}
  \op{add(\lit{absence}{R,T})} \la  &\brolit{\timesensor}{now}{T}, \brclit{ig}{focus}{R},\\
  &\naf\ \brclit{ig}{humanpos}{R},\\
  &\naf\ \brclit{ig}{absence}{R,T'},T'<T.
\end{align*}
as well as a bridge rule to forget the absence in a room if it is no longer necessary. There the delAll operator removes all occurrences of absence with respect to a given room $R$ from the KB of the context.
\begin{align*}
  &\op{delAll(\pred{absence},\term{R})} \la \brclit{ig}{humanpos}{R}.\\
  &\op{delAll(\pred{absence},\term{R})} \la \naf\ \brclit{ig}{focus}{R}.
\end{align*}
With those modifications it is  possible to generate an alert if Bob was too long away from the kitchen although the stove is active.
\end{example}
\section{Control of computation}\label{sec:control}
In this section we show how it is possible - at least to some extent - to control the effort spent on the computation of particular contexts. 
We introduce a specific control context $C_0$ which decides whether a context it controls should be idle for some time. An idle context just buffers sensor data it receives, but does not use the data for any other computations.

Let's illustrate this continuing the discussion of Sect.~\ref{sec:forget}. Assume there are $k$ different contexts for detecting potential emergencies as described earlier. The rMCS we are going to discuss is built on an architecture where each detector context $C_i$, $1 \leq i \leq k$ is connected via bridge rules with the control context. $C_0$ receives information about suspected emergencies and decides, based on this information, whether it is safe to let a context be idle for some time.

We now address the question what it means for a detector context to be idle. A detector context $C_i$ receives relevant observations 
to reason 
whether an emergency is suspected or confirmed. In case $C_i$ is idle, we cannot simply forget about new sensor information as it may become relevant later on, but we can buffer it so that it does not have an effect on the computation of a belief set, besides the fact that a buffered information shows up as an additional atom in the belief set which does not appear anywhere in the context's background knowledge.

To achieve this we have to modify $C_i$'s original bridge rules by adding, to the body of each rule, the context literal $\naf\ \brclit{0}{idle}{i}$. This means that the bridge rules behave exactly as before whenever the control context does not decide to let $C_i$ be idle.

For the case where $C_i$ is idle, i.e. where the belief set of $C_0$ contains $\lit{idle}{i}$, we just make sure that observations are buffered. This means that for each rule of the form $$\op{add(\pred{P},\term{T},\mathit{j_r})} \la \bropred{j_r}{P}, \brolit{\timesensor}{now}{T}$$ in the original set of bridge rules we add
$$\op{bf(\pred{P},\term{T},\mathit{j_r})} \la \bropred{j_r}{P}, \brolit{\timesensor}{now}{T}, \brclit{0}{idle}{I}.$$
The operation $\op{bf}$ just adds the atom $\lit{bf}{p,t,j_r}$ to the context (we assume here that the language of the context contains constructs of this form). As mentioned above, this information is not used anywhere in the rest of the context's KB, it just sits there for later use.

The only missing piece is a bridge rule bringing back information from the buffer when the context is no longer idle. This can be done using the bridge rule $\op{empty.buffer} \la \naf\ \brclit{0}{idle}{I}$. Whenever the management function has to execute this operation, it takes all information out of the buffer, checks whether it is still within the relevant time window, and if this is the case adds it to the KB, handling potential inconsistencies the way discussed in Sect.~\ref{sec:data}.

The control context uses formulas of the form $\lit{idle}{i,t}$ to express context $i$ is idle until time $t$. We intend here to give a proof of concept, not a sophisticated control method. For this reason we  simply assume the control context lets a detector context be idle for a specific constant time span $c$ whenever the detector does not suspect an emergency. This is achieved by the following bridge rule schemata:
\begin{tabbing}
xxx \= xxxxxxxxxxxxxxxxxx \= xx \= \kill
\> $\op{add(\lit{suspicion}{K})}$ \> $\la$ \> $\brclit{K}{susp}{E}$ \\
\>  $\op{add(\lit{idle}{K,T+c})}$ \>  $\la$ \> $\brolit{\timesensor}{now}{T}, \naf\ \brclit{0}{suspicion}{K},$ \\
\> \> \>  $ \naf\ \brclit{0}{idle}{K,T'}, T' < T+c$
\end{tabbing}

Provided information of the form $\lit{idle}{i,t}$ is kept until the actual time is $t+2$,  the last 2 conditions in the second rule schema guarantee that after being idle for period $c$ the context must check at least once whether some emergency is suspected. To avoid a context staying idle forever, we  assume the management function deletes information of this form whenever $t$ is smaller than the current time minus 1.
One more rule schema to make sure information about idle contexts is available in the form used by detector contexts:
$$\op{add(\lit{idle}{K})} \la \brolit{\timesensor}{now}{T}, \brclit{0}{idle}{K,T'}, T \leq T'.$$

\section{Complexity}
We want to analyze the complexity of queries on runs of rMCSs.
For simplicity we do not consider parametrized bridge rules here,
and assume that all knowledge bases in rMCSs are finite and
all management functions can be evaluated in polynomial time.
\begin{definition}\label{def:complDecProb}
The problem $\problemE$,
respectively $\problemA$,
is deciding whether for
a given rMCS $M$ with sensors $\Pi$, a context $C_i$ of $M$, a belief $b$ for $C_i$, and a finite sequence of $\Pi$-observations $O$
it holds that
$b\in S_i$  for some $\bstate^j=\tuple{S_1,\dots,S_n}$ ($0\leq j\leq n$) for some run, respectively all runs, $R = (\tuple{\bstate^0,\KBs^0}, \ldots, \tuple{\bstate^m,\KBs^m})$ of $M$ induced by $O$.
\end{definition}
As the complexity of an rMCS depends on that of its individual contexts
we introduce the notion of \define{context complexity}
along the lines of Eiter \etal~\cite{EiterFSW10}.
To do so, we need to focus on relevant parts of belief sets by means of projection.
Intuitively, among all beliefs, we only need to consider belief $b$ that we want to query and beliefs that
contribute to the application of bridge rules for deciding  $\problemE$ and $\problemA$.
Given $M$, $\Pi$, $C_i$, and $b$ as in Definition~\ref{def:complDecProb},
the \define{set of relevant beliefs} for a context $C_j$ of $M$ is given by
$RB_j(M,\brcpred{i}{b})=\set{b' \mid r\in br_j,\brc{h}{b'}\in\body{r}\lor \naf\ \brc{h}{b'}\in\body{r}}\cup\{b\mid i=j\}$.
A \define{projected belief state} for $M$ and $\brcpred{i}{b}$ is a tuple $S_{\mid M}^{\brcpred{i}{b}}=\tuple{S_1\cap RB_1(M,\brcpred{i}{b}),\dots,S_n\cap RB_n(M,\brcpred{i}{b})}$ where $\bstate=\tuple{S_1,\dots,S_n}$ is a belief state for $M$.
The \define{context complexity} of $C_j$ in $M$ \wrt $\brcpred{i}{b}$ for a fixed $\Pi$-observation $Obs$
is the complexity of deciding whether
for a given projected belief state $\bstate$ for $M$ and $\brcpred{i}{b}$,
there is some belief state $\bstate' = \tuple{S'_1, \ldots, S'_n}$ for $M$
with ${\bstate'}_{\mid M}^{\brcpred{i}{b}}=\bstate$ and $S'_j \in ACC_j(mng_j(\mi{app}_j(\bstate,Obs),kb_j))$ for all $1\leq j\leq n$.
The system's context complexity $\mathcal{CC}(M,\brcpred{i}{b})$ is a (smallest) upper bound for the context complexity classes of its contexts.
Our complexity results are summarized in Table~\ref{tab:complexity}.
\begin{table}
\vspace{-1.2mm}
\begin{center}
{\caption{Complexity of checking $\problemE$ and $\problemA$ (membership, completeness holds given hardness for $\mathcal{CC}(M,\brcpred{i}{b})$).}\label{tab:complexity}}
\vspace{-1.3mm}
$$
\begin{array}{|l|ll|}
\hline
\mathcal{CC}(M,\brcpred{i}{b})&\problemE&\problemA\\
\hline
\mathbf{P} & \mathbf{NP} & \mathbf{coNP}\\
\mathbf{\Sigma^{P}_{i}} (i\ge 2) & \mathbf{\Sigma^{P}_{i}} & \mathbf{\Pi^{P}_{i}}\\
\mathbf{PSPACE} & \mathbf{PSPACE} & \mathbf{PSPACE}\\
\hline
\end{array}
$$
\end{center}
\vspace{-1.8mm}
\end{table}

\noindent
Membership for $\problemE$: a non-deterministic Turing machine can guess
a projected belief state $\bstate^j=\tuple{S_1,\dots,S_n}$ for all $m$ observations in $O$ in polynomial time.
Then, iteratively for each of the consecutive observations $obs_j$,
first the context problem can be solved polynomially or using an oracle
(the guess of $\bstate^j$ and the oracle guess can be combined which explains that we stay on the same complexity level for
higher context complexity).
If the answer is 'yes', $\bstate^j$ is a projected equilibrium.
We can check whether $b\in S_i$, compute the updated knowledge bases
and continue the iteration until reaching the last observation.
The argument is similar for the co-problem of $\problemA$.
Hardness: holds by a reduction from deciding equilibrium existence for an MCS when $\mathcal{CC}(M,\brcpred{i}{b})$ is polynomial
and by a reduction from the context complexity problem for the other results.

Note that $\problemE$ and $\problemA$ are undecidable if we allow for infinite observations.
The reason is that rMCSs are expressive enough (even with very simple context logics) to simulate a Turing machine
such that deciding $\problemE$ or $\problemA$ for infinite runs solves the halting problem.

\section{Discussion}
In this paper we introduced reactive MCSs, an extension of managed MCSs for online reasoning, and showed how they allow us to handle typical problems arising in this area. Although we restricted our discussion to deterministic management functions,
two sources of non-determinism can be spotted by
the attentive reader.
On the one hand, we allow for semantics that return multiple belief sets
for the same knowledge base, and, on the other hand,
non-determinism can be introduced through  bridge rules.

The simplest example is guessing via positive support cycles, \egc using
bridge rules like
$
\begin{array}{l}
\op{add(\pred{a})}\la\brcpred{c}{a}\\
\end{array}
$
that allow (under the standard interpretation of $\op{add}$) for belief sets with and without formula~$\pred{a}$.
Multiple equilibria may lead to an exponential number of runs.
In practice, non-determinism will
have to be restricted.
A simple yet practical solution is to focus on a single run,
disregarding alternative equilibria.
Here, one might ask which is the best full equilibrium to proceed with.
In this respect, it makes sense to differentiate between non-deterministic contexts and
non-determinism due to bridge rules.
In the first case, it is reasonable to adopt the point of view of the answer-set programming (ASP) paradigm, \iec
the knowledge bases of a context can be seen as an encoding of a problem
such that the resulting belief sets correspond to the problem solutions.
Hence, as every belief set is a solution it does not matter which one to choose.
Thus, if the problem to be solved is an optimisation problem that has better and worse
solutions, this could be handled by choosing a context formalism able to express
preferences so that the semantics only returns sufficiently
good solutions.
For preferences between equilibria that depend on the belief sets of multiple contexts,
one cannot rely on intra-context preference resolution.
Here, we refer the reader to preference functions as proposed by Ellmauthaler~\cite{Ellmauthaler2013a}.
One might also adopt language constructs for expressing preferences in ASP
such as optimization statements~\cite{GKKOST10}
or weak constraints~\cite{BuccafurriLR97}.
Essentially, these assign a quality measure to an equilibrium.
With such additional quality measures at hand, the best equilibrium can be chosen for the run.

As to related work, there is quite some literature on MCSs by now, for an overview see \cite{BrewkaEF11}.
Recently an interesting approach to belief change in MCSs has been proposed~\cite{Wang2013}.
Other related work concerns stream reasoning in ASP~\cite{Gebser2012} and in databases: a continuous version of SPARQL~\cite{Barbieri2010} exists, and logical considerations about continuous query languages~\cite{Zaniolo2012} were investigated.
Kowalski's logic-based framework for computing~\cite{Kowalski2013} is an approach which utilizes first order logic and concepts of the situation- and event-calculus in response to observations.
Updates on knowledge-bases, based upon the outcome of a given semantics where also facilitated for other formalisms, like logic programming in general. There the iterative approaches of {\sc epi}~\cite{Eiter2001} and {\sc evolp}~\cite{Alferes2002} are the most prominent.
Note that none of these related approaches combines a solution to both knowledge integration and online reasoning, as we do.

The idea of updates to the knowledge-base was also formalised for database systems~\cite{Baral1997}.

For a related alternative approach using an operator for directly manipulating KBs without contributing to the current equilibrium,
we refer to the work by Gon\c{c}alves, Knorr, and Leite~\cite{GKL2014}.

\end{document}